\documentclass{article}

\PassOptionsToPackage{square,numbers}{natbib}

\usepackage[final]{neurips_2021_ml4ps}

\usepackage[utf8]{inputenc} 
\usepackage[T1]{fontenc}    
\usepackage{hyperref}       
\usepackage{url}            
\usepackage{booktabs}       
\usepackage{amsfonts}       
\usepackage{nicefrac}       
\usepackage{microtype}      
\usepackage[dvipsnames]{xcolor}         
\usepackage{multirow}
\usepackage{hhline}

\title{Uncertainty quantification for ptychography using normalizing flows}

\author{%
  Agnimitra Dasgupta \\
  Sonny Astani Department of Civil and Environmental Engineering\\
  University of Southern California\\
  Los Angeles, CA 90089 \\
  \texttt{adasgupt@usc.edu} \\
  \And
  Zichao Wendy Di \\
  Mathematics and Computer Science Division \\
  Argonne National Laboratory \\
  \texttt{wendydi@anl.gov}
}

 \usepackage{amsfonts}
\usepackage{graphicx}
\usepackage{epstopdf}
\usepackage{mathrsfs}
\usepackage{hyperref}
\usepackage{amsmath}
\usepackage{amsfonts}
\usepackage{nicefrac}
\usepackage{latexsym}
\usepackage{comment}
\usepackage{color}
\usepackage{listings}
\usepackage{placeins}
\usepackage{makecell}
\usepackage{tikz}
\usepackage{subcaption}
\usepackage[labelfont=bf]{caption}
\usepackage{cleveref}
\usepackage{bm}
\usepackage{wrapfig}
\usepackage{soul}

\usepackage{enumitem}
\setlist[enumerate]{leftmargin=.5in}
\setlist[itemize]{leftmargin=0.25in}

\newcommand{\supth}[1]{\ensuremath{{#1}^{\text{th}}}}

\usepackage{amsopn}
\graphicspath{{figures/}}

\newcommand{\ds}{\displaystyle}
\newcommand{\Fcurly}{\mathcal{F}}

\newcommand{\Rcurly}{\mathcal{R}}

\newcommand{\bfv}{ {\bf{v}}}

\newcommand{\bfP}{{\bm P}}

\newcommand{\bfd}{{\bm d}}

\newcommand{\bfw}{{\bm w}}

\newcommand{\bfu}{{\bm u}}
\newcommand{\bfz}{{\bm z}}

\newcommand{\Lcurly}{\mathcal{L}}

\newcommand{\bfepsilon}{\boldsymbol \epsilon}

\usepackage{mathtools}
\usepackage{url}

\usepackage{mathtools}

\usepackage{array}

\newlength{\myrowheight}
\newcommand{\WDNote}[1]           
{\textcolor{red}{#1}\marginpar{\textcolor{red}{WD $\longleftarrow$}}}

\begin{document}

\maketitle

\begin{abstract}
Ptychography, as an essential tool for high-resolution and nondestructive material characterization, presents a challenging large-scale nonlinear and non-convex inverse problem; however, its intrinsic photon statistics create clear opportunities for statistical-based deep learning approaches to tackle these challenges, which has been underexplored.   
In this work, we explore normalizing flows to obtain a surrogate for the high-dimensional posterior, which also enables the characterization of the uncertainty associated with the reconstruction: an extremely desirable capability when judging the reconstruction quality in the absence of ground truth, spotting spurious artifacts and guiding future experiments using the returned uncertainty patterns. 
We demonstrate the performance of the proposed method on a synthetic sample with added noise and in  various physical experimental settings. 
\end{abstract}

\section{Introduction}

Ptychography is a special type of coherent diffraction imaging technique, wherein an object is scanned at a series of overlapping regions using a coherent beam of incident waves and reconstructed from the series of diffraction patterns~\cite{rodenburg2008ptychography, enders2016computational}. 
As a lensless imaging technique, ptychography has become an essential tool for high-resolution and nondestructive material characterization, with state-of-the-art techniques reaching below 10 nm in spatial resolution \cite{pfeiffer2018x}, across a wide variety of disciplines including materials science  and biology; see \cite{konda2020fourier} and references therein. However, the aforementioned advantages come at the expense of having to tackle a nonlinear and non-convex (i.e., ill-posed as presenting nonunique local minima) inverse problem. 
Traditionally, ptychographic reconstruction has been solved iteratively using deterministic optimization-based approaches. Due to the ill-posed nature, the solution quality returned by traditional approaches can contain artifacts and  is usually highly sensitive to initialization, model accuracy, and the underlying optimization solver used, especially in the presence of noise~\cite{chang2019advanced}. Therefore, in the absence of ground truth, robust characterization (i.e., uncertainty quantification) of the different solutions, and any artifacts therein, is highly desirable. 

End-to-end approaches using deep learning have become popular in ptychographic reconstruction \cite{cherukara2020ai} and other phase imaging methods \cite{sinha2017lensless,li2018imaging,rivenson2018phase,ongie2020deep} with the focus to obtain a reconstruction efficiently. 
There has been limited work demonstrating uncertainty quantification on phase imaging using Bayesian convolution neural networks (e.g., \cite{xue2019reliable,wei2020uncertainty}); albeit in a supervised learning setting such that any uncertainty descriptions are conditioned on the training set. \citet{wei2020cramer} used the Cram\'er-Rao bound to derive analytical lower bound for uncertainty in ptyhcographic reconstruction. Their approach, however, remains limited to Poisson noise models and will be challenging to extend to more complicated models.

One potential candidate for a systematic and robust uncertainty quantification is using Bayesian approaches to obtain the posterior distribution and associated statistics. Typically, for nonlinear inverse problems, the posterior is approximated using samples generated with the help of Markov-chain Monte-Carlo (MCMC) methods~\cite{bardsley2012mcmc}. However, MCMC can be inefficient for high dimensional problems such as ptychography; slower mixing of the Markov chains leading to drastically increased computational costs~\cite{cui2014likelihood}. Normalizing flows (NFs) are a class of generative models that can transform a probability density, usually simple and easy to sample, into another more complicated probability density through a sequence of invertible mappings \cite{rezende2015variational,kobyzev2020normalizing,papamakarios2021normalizing}. One benefit of flow-based generative models, compared to other types of generative models, is its capability of computing likelihood exactly and efficiently~\cite{kingma2018glow}. Therefore, NFs have been popular choices to approximate complicated posterior densities. For example, in computer scinece, NFs have been successfully applied in various problems including computer vision, speech synthesis, etc. \cite{kingma2018glow,ardizzone2019guided,ziegler2019latent,prenger2019waveglow,kumar2019videoflow}, and recently been explored for various inverse problems in natural science ranging from cosmology to medical imaging \cite{zhu2019physics,mo2019deep,padmanabha2021solving,siahkoohi2021learning,williams2021nested,sun2020deep,albergo2019flow,ardizzone2018analyzing}. 
 
 Inspired by the unsupervised deep probabilistic imaging framework proposed by \citet{sun2020deep},  in this work, we explore the applicability of NFs for the task of reconstruction and uncertainty quantification on ptychography. More specifically, we construct a surrogate for the posterior density of the reconstructed object using flow-based generative models. Samples generated using such \emph{trained} surrogates can be used to obtain reconstructions of the object and estimate standard deviations of the reconstructions. In this work, we use the standard deviation of the reconstructed phase and magnitude as the metric for uncertainty quantification, and demonstrate that the reconstruction quality obtained by the proposed NFs is comparable to an advanced ptychographic reconstruction methods "rPIE" \cite{maiden2017further}
under various scan settings that reflect varied degree of ill-posedness. Furthermore, compared to traditional methods, our approach presents a unique capability of quantifying the solution quality in the absence of ground truth.

\section{Methods}
\paragraph{Forward model:} Let $\bfz\in \mathbb{C}^{n^2}$ be a complex variable representing the object of interest, and $\bfd_j \in \mathbb{R}^{m^2}$ be the observed data (or intensities) measured from the $j^{th}$ scanning position, where $n^2$ and $m^2$ are the dimensions of the vectorized object and data, respectively. A ptychography experiment is modeled by
\begin{align} \label{eq:forwardProblem}
    \bfd_j = |\Fcurly(\bfP_j\bfz)|^2 + \bfepsilon_j, \quad j=1,\ldots,N,
\end{align}
where $N$ is the total number of scanning positions, $\Fcurly\colon \mathbb{C}^{m^2} \mapsto \mathbb{C}^{m^2}$ is the two-dimensional discrete Fourier operator, $\bfP_j \in \mathbb{C}^{m^2 \times n^2}$ is the given illumination probe at the $j^{th}$ step, and $\bfepsilon_j \in \mathbb{R}^{n^2}$ is the noise corresponding to the $j^{th}$ measurement. 
Then we have the ptychographic forward model as $f_j(\bfz)=|\Fcurly(\bfP_j\bfz)|^2$ for the $j^{th}$ scan, which is nonlinear and non-convex, and $f(\bfz) = \{ f_j(\bfz) \}_{j=1}^{N}$. 



\paragraph{Approximating the posterior distribution using normalizing flows:}
A surrogate $q(\bfz)$ for the posterior density $p(\bfz|\bfd)$ can be constructed by letting $\bfz = \mathcal{G}_\theta(\bfw)$ where $\bfw \sim \mathcal{N}(0, \mathcal{I}_{2n^2})$\footnote{In this work, due to the presence of complex number, the actual output from the NF is $2n^2$ real numbers which is divided into two chunks that are used separately as the real and imaginary part of each pixel to form $\bfz$.}. $\mathcal{G}$ is a generative model (in this case an NF), and $\theta$ is its parameters. For $q(\bfz)$ to be a valid probability distribution, the mapping $\mathcal{G}$ must be bijective, which gives us
\begin{equation*}
    \log q(\bfz) = \log p(\bfw) - \log \left| \det \frac{\partial \mathcal{G}_\theta(\bfw) }{ \partial \bfw } \right|,
\end{equation*}
due to the change of variables rule. 
Therefore, we use invertible neural networks (INNs) to form $\mathcal{G}_\theta$ to ensure its bijectiveness. The parameters $\theta$ can be learned by minimizng the Kullback-Liebler  distance between $q(\bfz)$ and $p(\bfz|\bfd)$ which leads to the following minimization problem \cite{sun2020deep}
\begin{equation}\label{eq:objectivefunction}
    \theta^\ast = \operatorname*{arg\,min}_{\theta} \sum_{k=1}^{B} \Lcurly(\bfd | f(\mathcal{G}_\theta(\bfw_k))) + \lambda \Rcurly(\mathcal{G}_\theta(\bfw_k)) - \log \left| \det \frac{\partial \mathcal{G}_{\theta}(\bfw) }{ \partial \bfw} \right|_{\bfw = \bfw_k},
\end{equation}
where $\bfw_k$ is the \supth{k} sample,  B is the batch size for training, and  $\Lcurly$ and $\Rcurly$ are the data misfit and regularization terms, respectively, such that \( p(\bfz|\bfd) \propto \exp \left( - \Lcurly(\bfd|\bfz) - \lambda\Rcurly(\bfz) \right) \) where $\bfz = \mathcal{G}_\theta(\bfw)$. $\lambda$ is the regularization weight which requires tuning to be relatively optimal. Additionally, NFs do not require training data separately \protect\cite{sun2020deep}. Instead the cost function (2) can be optimized to learn the weights and concurrently monitored to tune any hyperparameters. 

There are various ways to model the data misfit in ptychography given Eqn.~\eqref{eq:forwardProblem}. Given $\sigma_j$ as the assumed variance of $\bfepsilon_j$, we focus on the amplitude-based error metric which is
\begin{equation}\label{eq:misfit1}
  \ds \Lcurly\left(\bfd_j|f_j(\bfz)\right) = \frac{1}{2\sigma_j^2} \left\| \left|  \mathcal{F}(\bfP_j \bfz)\right| - \sqrt{\bfd_j} \right\|_2^2, \quad \text{s.t} \;\;\Lcurly\left(\bfd|f(\bfz)\right) = \sum_{j=1}^{N} \Lcurly\left(\bfd_j|f_j(\bfz)\right),
\end{equation}
due to its stability, as compared to other options such as the intensity Gaussian metric \cite{fung2020multigrid}. 

\paragraph{Constraints:} The only constraint we enforce is a $[0,1]$ box constraint on the magnitude (i.e., $|\bfz|$) of an object, as a common assumption corresponding to the absorption capability of material \cite{swinehart1962beer}. In our implementation, we constrain the magnitude, which is formed by querying the surrogate, by normalizing it via $\bfz \leftarrow \bfz' / || \bfz' ||_\infty $ where $\bfz' = \mathcal{G}_\theta(\bfw)$, before evaluating the cost function~\eqref{eq:objectivefunction}.

\begin{wrapfigure}[8]{r}{0.6\textwidth}
    \centering
   \vspace{-1.1\baselineskip}
    \includegraphics[width=0.6\textwidth]{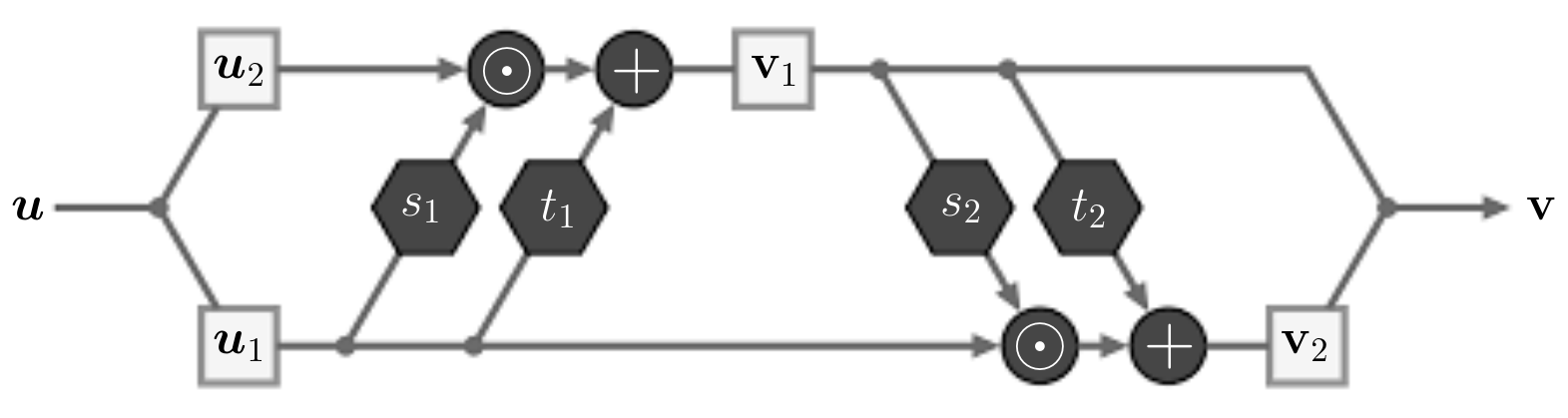}
    \caption{Schematic of a coupling layer}
    \label{fig:coupling}
\end{wrapfigure}

\paragraph{Coupling layers:}
We construct NFs using INNs that are based on the coupling architecture proposed by \citet{dinh2014nice} and later refined in \cite{dinh2016density}. The coupling layer (see Fig.~\ref{fig:coupling}) works as follows: an input $\bfu$ is divided into two parts $\bfu_1$ and $\bfu_2$ which are transformed into corresponding parts $\bfv_1$ and $\bfv_2$ of the output $\bfv$ using affine transformations as follows 
\begin{equation*}
      \bfv_1 = \bfu_2 \odot \exp[s_1(\bfu_1)] + t_1(\bfu_1)\,  \rightarrow \,\bfv_2 = \bfu_1 \odot \exp [s_2(\bfv_1)] + t_2(\bfv_1)\\
\end{equation*}

    where $s_1, s_2, t_1, t_2$ are scale and shift operators modeled using standard neural networks (detailed in Sec.~\ref{sec:numericalresults}) 
    A deep INN can then be constructed by composing K such coupling layers. The overall representation power of the deep INN can be enhanced by permuting the inputs to each coupling layer \cite{ardizzone2018analyzing,kingma2018glow}. The use of coupling layer constrains the Jacobian to be upper triangular; meaning \( \det \nicefrac{\partial \mathcal{G}(\bfw)}{\partial\bfw} \) in Eqn.~\ref{eq:objectivefunction} can be computed efficiently~\cite{dinh2014nice}.

\section{Numerical Results}\label{sec:numericalresults}
\begin{wrapfigure}[8]{r}{0.35\textwidth}
 \vspace{-4\baselineskip}
  \begin{center}
    \includegraphics[height=2.5cm]{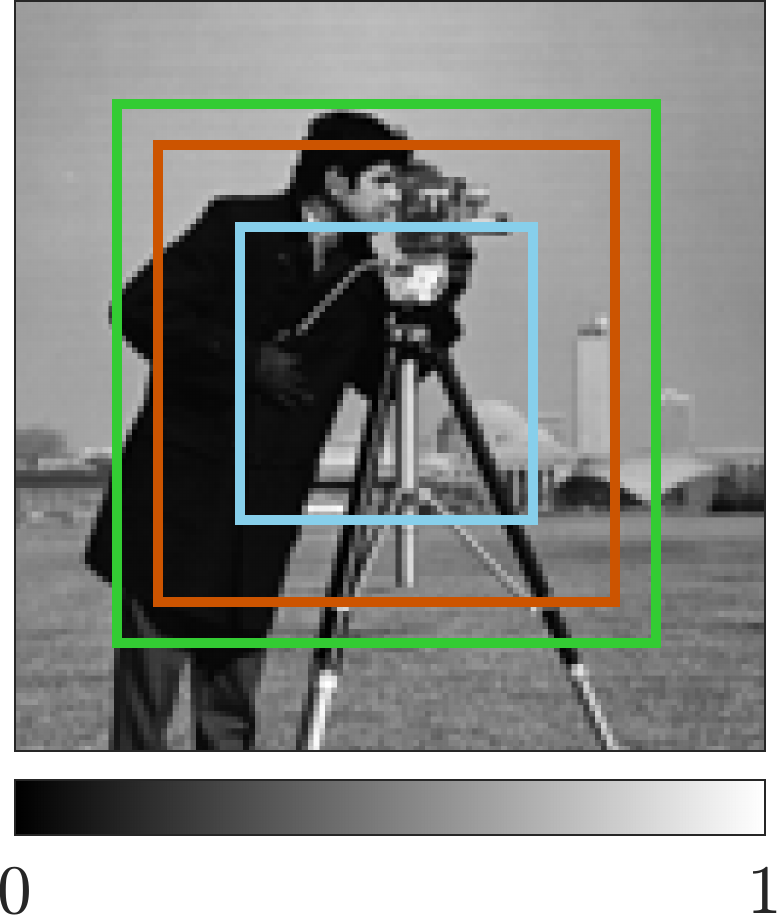}
    \includegraphics[height=2.5cm]{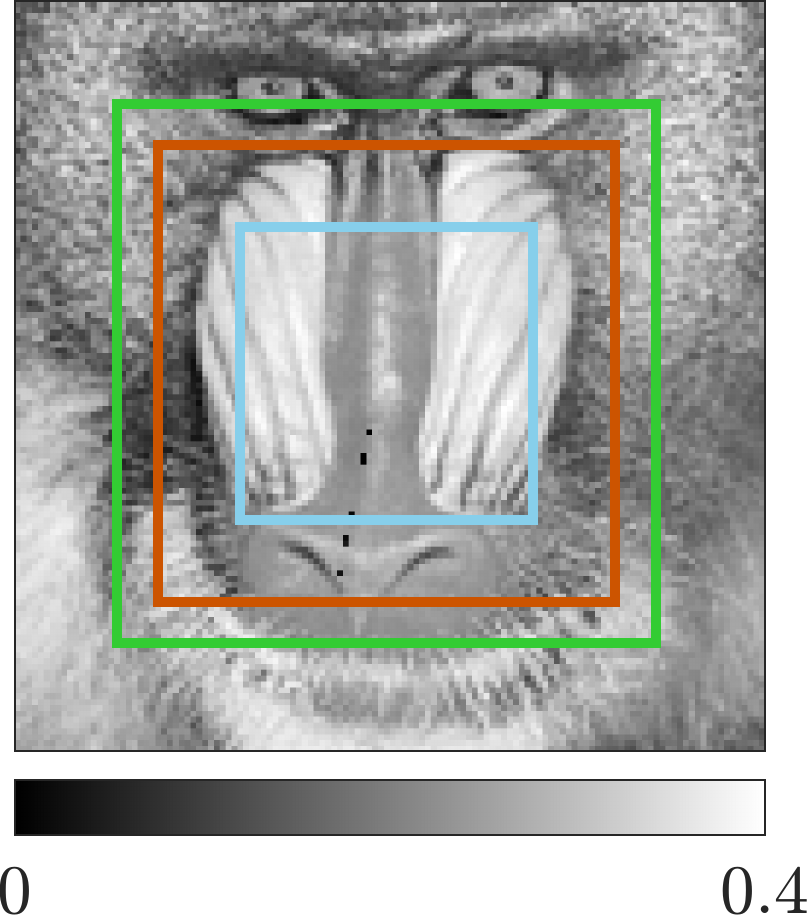}
  \end{center}
  \vspace{-0.5em}
  \caption{Ground truth object: magnitude (left) and phase (right).}
  \label{fig:sample}
\end{wrapfigure}
We demonstrate the proposed approach on a synthetic sample (see Fig.~\ref{fig:sample}) having the cameraman and baboon images as the magnitude ($|\bfz|$) and phase ($\operatorname{arg}(\bfz)$) components of $\bfz$, respectively. The object is scanned $8$ times along each axis (i.e.,  $N = 64$) using a probe of size $m=36$. 
We investigate three different scanning settings corresponding to three different strides of the probe. Associated with each setting is a field of view (FOV) corresponding to the different region of the image that being illuminated (i.e., different $n$, see  color-coded boundaries in Fig.~\ref{fig:sample}), however, with the same number of data, i.e., $\bfd$ is of dimension $64 \times 36 \times 36$ for all three settings. 
Furthermore, we add 1\% Gaussian noise to the measurements to reflect practical scenarios. The sensor noise can be modeled as Gaussian when the photon counts are large due to the fact that the limiting distribution of the Poisson distribution is Gaussian.
\begin{table}[h]
    \setlength\tabcolsep{0pt}
    \setlength\extrarowheight{2pt}
    \caption{Different scan settings and reconstruction results}
    \label{tab:expsettings}
    \begin{tabular*}{\textwidth}{@{\extracolsep{\fill}}*{9}{c}}
        \hline\hline
        \multirow{2}{*}{\makecell{Scan\\ Setting}} & \multirow{2}{*}{\makecell{FOV \\ $n$}} & \multirow{2}{*}{\makecell{Overlap\\ Ratio}} & \multicolumn{3}{c}{Recon. Mag. PSNR} & \multicolumn{3}{c}{Recon. Phase SSIM} \\\cline{4-6} \cline{7-9}
        &&&NF Mode 1&NF Mode 2&rPIE&NF Mode 1&NF Mode 2&rPIE \\\hline
        \textcolor{SkyBlue}{$S1$} & $50$ & 0.94 & 26.69 & 26.64 & 26.82 & 0.48 & 0.54 & 0.25\\
        \textcolor{BurntOrange}{$S2$} & $78$ & 0.83 & 22.77 & 23.66 & 23.08 & 0.20 & 0.39 & 0.18\\
        \textcolor{OliveGreen}{$S3$} & $92$ & 0.78 & 21.20 & 22.69 & 22.32 & 0.50 & 0.44 & 0.16\\
        \hline
   \end{tabular*}
\end{table}

For all the results shown herein, we use deep INNs with 16 blocks. Each block consists of two affine coupling layers, as shown in Fig.~\ref{fig:coupling}, and each coupling layer is preceded by an \emph{activation normalization} unit \cite{kingma2018glow}. The dimensions are permuted and divided into two equal parts at the beginning of each block. The scale and shift transformations are modeled using dense neural networks with two $n^2/4$-dimensional hidden layers. Leaky RELU activation is used and batch normalization is also applied. The INNs are trained for $2 \times 10^4$ epochs with a learning rate of $5 \times 10^{-6}$ and a batch size of $64$. Given $\Lcurly$ as Eqn.~\eqref{eq:misfit1}, $\Rcurly$ corresponds to a total variation prior \cite{rudin1992nonlinear} with $\lambda = 0.01$. 
Following \cite{fung2020multigrid}, the reconstructed magnitudes are compared using PSNR whereas the reconstructed phases are compared using SSIM after the predicted phase is linearly mapped to the range ground truth region $[0, 0.4]$. Table~\ref{tab:expsettings} details the scanning settings with the corresponding reconstruction results using NFs and their comparison against rPIE. All experiments were performed on a computer with a 48GB NVIDIA Quadro RTX 8000 GPU and 64GB of RAM. 



\begin{wrapfigure}[11]{r}{0.5\textwidth}
    \vspace{-1\baselineskip}
    \includegraphics[width=0.5\textwidth]{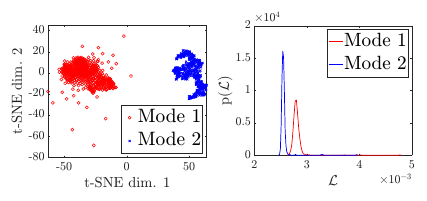}
    \vspace{-1\baselineskip}
    \caption{Left: bi-modality revealed by t-SNE; right: distribution of $\Lcurly$ within each mode with mode 1 having higher bias.}
    \label{fig:postt-SNES2}
\end{wrapfigure}

\begin{wrapfigure}[13]{l}{0.6\textwidth}
    \vspace{-1\baselineskip}
    \centering
    \begin{tikzpicture}
        \node at (0,0) {\includegraphics[width=0.60\textwidth]{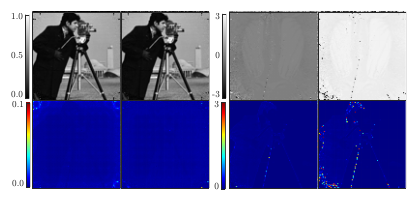}};
        \node at (-2.7,2.0) {Mode 1};
        \node at (-0.9,2.0) {Mode 2};
        \node at (1.2,2.0) {Mode 1};
        \node at (3.1,2.0) {Mode 2};
        \node[align=center, rotate=90] at (-4.1,1.0) {Mean};
        \node[align=center, rotate=90] at (-4.1,-0.8) {SD};
    \end{tikzpicture}
    \vspace{-2em}
    \caption{Posterior mean and  standard deviation (SD) of the reconstructed magnitude and phase for both modes.}
    \label{fig:postt-SNES1}
\end{wrapfigure}
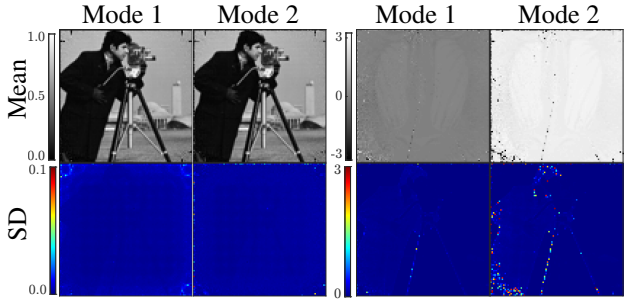

We first focus on setting $S3$, and perform an analysis of the samples drawn from the generative model using the t-distributed stochastic neighbor embedding (t-SNE) \cite{van2008visualizing}, which reveals bi-modality in the posterior (see Fig.~\ref{fig:postt-SNES2}(left)). These modes reflect the classical \textit{bias-variance trade-off} of estimators. Figure~\ref{fig:postt-SNES2}(right) shows that mode 1 has higher bias than mode 2 as evidenced by the slightly right-shifted distribution of $\mathcal{L}$ for mode 1. However, as seen later in Fig.~\ref{fig:postt-SNES1}, mode 1 has lower variance particularly for the phase reconstruction. 

In Fig.~\ref{fig:postt-SNES1}, we show the posterior mean and its associated uncertainty of the two modes ( labeled as mode 1 and 2, respectively).  
Both modes yield equally good reconstructions of the object magnitude; however, the phase reconstructions vary across the modes (see Table~\ref{tab:expsettings}).  In particular, as expected, the edges (particularly corners) of the FOV are reconstructed with higher standard deviations due to the limited overlap of the scans near the edges.

\vspace{\baselineskip}

\begin{wrapfigure}[10]{r}{0.4\textwidth}
    \vspace{-2\baselineskip}
    \centering
    \includegraphics[width=0.4\textwidth]{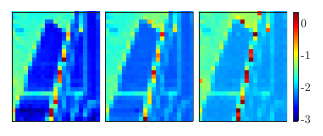}
    \vspace{-1.5em}
    \caption{Logarithm (base 10) of standard deviation of phase reconstruction in mode 2 for $S1$, $S2$ and $S3$ over common regions.}
    \label{fig:STDsettings}
\end{wrapfigure}
Moreover, the artifacts in the phase reconstruction (seen as the tripod leg), as a consequence of the cross talks between magnitude and phase, show higher standard deviations. This is an extremely useful result in order to judge the reconstruction quality in practice since the ground truth is often unavailable in real applications. 

We further quantify the reconstruction qualities caused by different scanning settings. As the settings move from $S1$ to $S3$, reconstruction task potentially becomes harder due to the less overlap. Consistently, the standard deviation increases over common areas of the three FOVs as the overlap ratio reduces, as shown in Fig.~\ref{fig:STDsettings}. The lower SSIM of the reconstructed phases in setting S2 as compared to S3 in \protect\Cref{tab:expsettings} may be an artefact that indicates the need for better hyperparameter tuning or longer training of the INNs.


\section{Conclusions}


We demonstrate that normalizing flows can be used as a framework for the statistical analysis of inverse problems arising in ptychography on a synthetic sample. First, the uncertainty quantification returned by the proposed NFs shows that the reconstruction quality  deteriorates as the overlap ratio reduces. This is expected since the overlap ratio is a measure of oversampling, less the extent of oversampling leads to greater the ill-posedness of the inverse problem. Second, with phase reconstruction as the main goal of phase retrievel in general, the proposed NF outperforms traditional method by providing better phase reconstruction. Overall, the main contribution of the proposed NF is its capability of quantifying uncertainties associated with the reconstruction, which is essential in judging the solution quality in the absence of ground truth. Future direction includes a careful prior design to provide more guidance on the mode selection. The application of the proposed methodology on a larger object with more data will be explored in a future work.

\begin{ack}
The first author acknowledges the support of the Provost Fellowship from the University of Southern California. 
\end{ack}

\newpage

{\small
\bibliographystyle{abbrvnat}
\bibliography{biblio} 
}

\newpage

\normalsize

\end{document}